\documentclass[10pt, a4paper]{article}

\usepackage{lrec}
\usepackage{multibib}
\newcites{languageresource}{Language Resources}
\usepackage{graphicx}
\usepackage{tabularx}
\usepackage{soul}
\usepackage[T1]{fontenc}
\usepackage{epstopdf}

\usepackage{hyperref}
\usepackage{xstring}

\usepackage[utf8]{inputenc} 

\title{Expanding Abbreviations in a Strongly Inflected Language: Are Morphosyntactic Tags Sufficient?}
\name{Piotr \.Zelasko}

\address{
  AGH University of Science and Technology, Krak\'ow, \\
  www.dsp.agh.edu.pl, \\
  pzelasko@agh.edu.pl
  }

\abstract{
  In this paper, the problem of recovery of morphological information lost in abbreviated forms is addressed with a focus on highly inflected languages. Evidence is presented that the correct inflected form of an expanded abbreviation can in many cases be deduced solely from the morphosyntactic tags of the context. The prediction model is a deep bidirectional LSTM network with tag embedding. The training and evaluation data are gathered by finding the words which could have been abbreviated and using their corresponding morphosyntactic tags as the labels, while the tags of the context words are used as the input features for classification. The network is trained on over 10 million words from the Polish Sejm Corpus and achieves 74.2\% prediction accuracy on a smaller, but more general National Corpus of Polish. The analysis of errors suggests that performance in this task may improve if some prior knowledge about the abbreviated word is incorporated into the model. \\ \newline \Keywords{abbreviation expansion, morphosyntactic tags, strongly inflected langauges}
}

\begin{document}

\maketitleabstract

\section{Introduction}
\label{sec:introduction}

In this paper, we address the problem of recovery of morphological information lost in abbreviated forms, focusing on highly inflected languages. Significance of the problem can be shown in the domain of text normalization, especially in cases of expanding numerals and abbreviations to their full forms. Although in English the translation of token "14" to "fourteen", or token "yr." to "year" appears trivial, it is not so in a highly inflected language. For example in Polish, the abbreviation "r." could be expanded into one of \{\textit{rok}, \textit{roku}, \textit{rokowi}, \textit{rokiem}\}, all of which mean "year" in English, but are inflected differently, based on the context in which they are used. Correct choice of the grammatical number and case is essential towards successful text normalization, which is an indispensable part of any text-to-speech (TTS) system \citelanguageresource{sproat2016rnn}, and can also be useful in text preprocessing for language model (LM) training \cite{pakhomov2002semi}.

Under the assumption that an abbreviation is not ambiguous, the conversion of a token into its abbreviated form can be seen as a removal of morphological information from the token. Nonetheless, a person (e.g. a native speaker) reading the text is able to deduce the information missing from the abbrevation by using the context. Furthermore, we suspect that it is possible to abstract away from concrete words and infer the missing morphological information based only on morphological and syntactic properties of the context. Consequently, we formulate a hypothesis, that based purely on the morphosyntactic information present in other words in the sentence (i.e. the context), the correct morphosyntactic tag can be inferred for the abbreviated word in a highly inflected language. 

Given recent major achievements of deep neural networks (DNN) in Natural Language Processing (NLP) tasks \cite{sundermeyer2012lstm} \cite{huang2015bidirectional}, we apply a bidirectional recurrent neural network (RNN) \cite{graves2005framewise} based on Long Short-Term Memory (LSTM) cells \cite{hochreiter1997long} to the task of missing morphosyntactic tag prediction. The network is trained on a large collection of Polish texts analysed with a morphosyntactic tagger and evaluated on a manually annotated corpus in order to verify the hypothesis. To the best of our knowledge, no previous attempt has been made in order to infer the correct form of an abbreviated word with an RNN, while using solely morphosyntactic information of its context within the sentence.

In section \ref{sec:related_work} we present previous relevant work in the area of text normalization. Section \ref{sec:methods} describes the problem of inflected expansion of abbreviations and the DNN architecture used to solve it. The data sets, experimental setting and results are shown in section \ref{sec:experiment} Finally, we conclude our work in section \ref{sec:conclusions}

\section{Related Work}
\label{sec:related_work}

Although there have been several studies regarding text normalization, it can be safely stated that it did not receive as much attention as some other areas of NLP. In \cite{pakhomov2002semi}, the author employed a Maximum Entropy (ME) model in order to disambiguate and expand the abbreviations present in medical documents. In order to provide the labels for supervised training of the model, sentences containing non-abbreviated forms were identified and treated as if the term had been abbreviated. The model was then trained to recognize correct expansion based on a context of 7 words on each side of the abbreviation. Even though in our work we adopted similar approach to training data collection, we use the morphosyntactic tags instead of words and utilize the full sentence instead of a fixed context window. Furthermore, our goal is to infer morphological information and not to disambiguate the abbreviation.

Similar efforts have been undertaken in the task of numerals normalization in the Russian language, which is also highly inflected. In \cite{sproat2010lightly}, the author had investigated the viability of n-gram, perceptron and decision list models and found that a trigram model slightly outperforms the others. He also suggested further investigation into application of more complex discriminative models.

A different branch of text normalization is focused on abbreviation discovery. One approach includes a character-level Conditional Random Field (CRF) model used to generate most likely deletion-based abbreviations for words, which are then reverse-mapped and disambiguated using an additional language model \cite{pennell2011toward}. In \cite{roark2014hippocratic}, the authors identified abbreviations by noticing that in a large data set, they appear in the same contexts as the non-abbreviated words. To expand them, they perform a fusion of N-gram and Support Vector Machine (SVM) models which is characterized by a low false alarm rate. 

Recently, there has been growing interest in text normalization seen as a machine translation task - in order to foster more research in this direction, a challenge with an open data set has been published \citelanguageresource{sproat2016rnn}. Regardless of promising results, the authors observed that the RNN tend to fail in some cases in quite a weird manner - such as translating abbreviated \textit{hours} as \textit{gigabytes}. Our approach differs from the one presented in \citelanguageresource{sproat2016rnn} in the way the RNN is used - instead of constructing a sequence to sequence (seq2seq) model, we use a many-to-one mapping in our network in order to identify only one missing piece of information at a time.

\section{Methods}
\label{sec:methods}

Let us consider a sentence $W=\{w_i\}$, where $w_i$ is the i-th word of the sentence. Each word (lemma) $w_i$ can be described as a combination of a basic form (lexeme) $l_i$ and morphosyntactic information $m_i$, expressed as a set of particular morphosyntactic tags. These tags inform about both morphological and syntactical properties of the word, such as grammatical class, number, case, gender, etc. 

\begin{figure}[ht]
\centering
\includegraphics[width=\columnwidth]{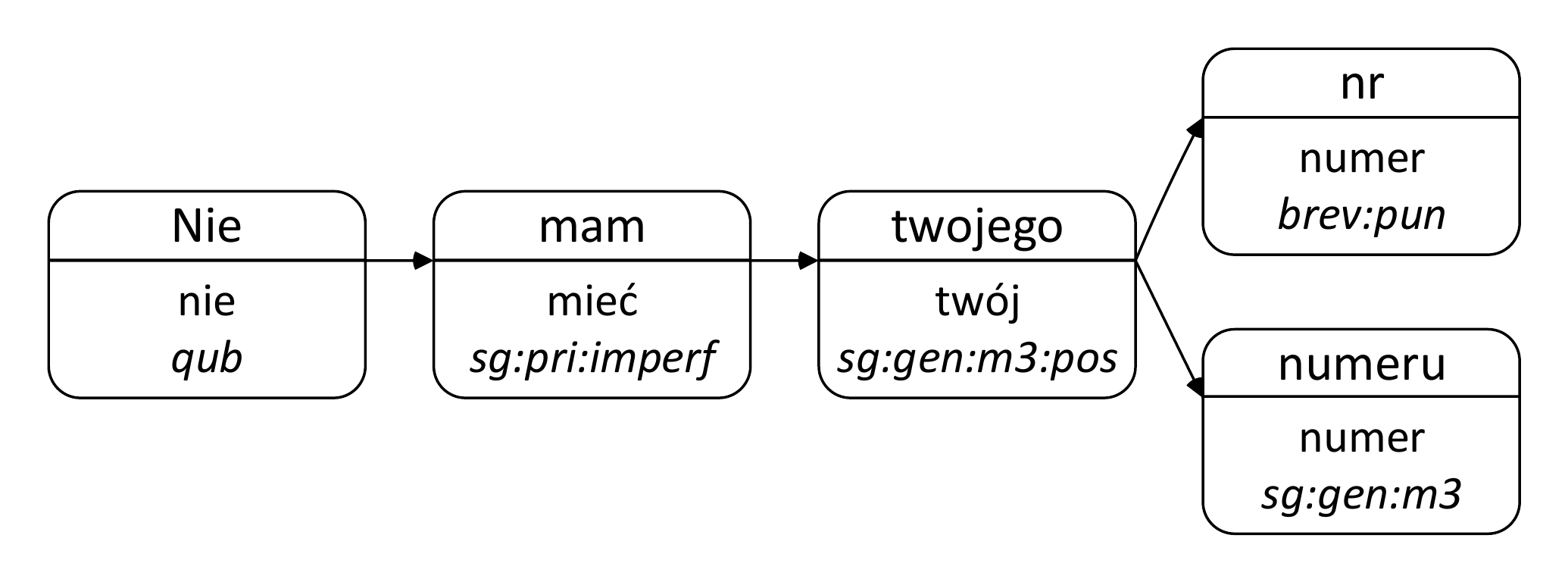}
\caption{A comparison of abbreviated and full form of the lemma \textit{numeru} in Polish equivalent of English sentence \textit{"I don't have your number"}. 
The upper block contains lemmas and the lower block their corresponding lexemes and morphological tags.}
\label{fig:sentence_example}
\end{figure}

In a sentence without any abbreviations, each lemma explicitly presents its morphosyntactic information through inflectional morphemes, which are composed together with the lexeme (e.g. lexeme \textit{pasture} composed with morpheme \textit{s} results in lemma \textit{pastures}, which informs that the grammatical number is plural). However, this is not the case in abbreviated words, which are not subject to inflection, regardless of their context. Nonetheless, the obfuscated morphosyntactic properties of an abbreviation can usually be deduced by a native speaker of an inflectional language through analysis of the context. Consequently, it can be said that abbreviations, presented within a context, contain \textit{implicit} morphosyntactic information. This phenomenon is illustrated in a sentence\footnote{
The usage of this abbreviated form in this context is not typical in Polish writing, but may be representative of an abbreviation used in a short message sent to a peer.}
in figure \ref{fig:sentence_example}, where the abbreviated form is shown to have unknown morphological properties (i.e. a tag \textit{brev:pun}, which indicates an abbreviation followed by punctuation), in contrast to the full form.

In order to predict this implicit morphosyntactic information, we propose the application of an RNN - in particular, a bidirectional LSTM network - which analyses the whole sentence.

The input of the network is a sequence of morphosyntactic tags $m_i$, each one corresponding to the word $w_i$ from the original sentence $W$. The unknown tag of the abbreviation is substituted by a special token, which indicates that this tag is to be inferred. The input layer is then connected to an embedding layer, which casts the tags into a low-dimensional real-valued vector, similarly as it is done in case of regular words in modern neural language models \cite{mikolov2013distributed}.

\begin{figure}[ht]
\centering
\includegraphics[width=\columnwidth]{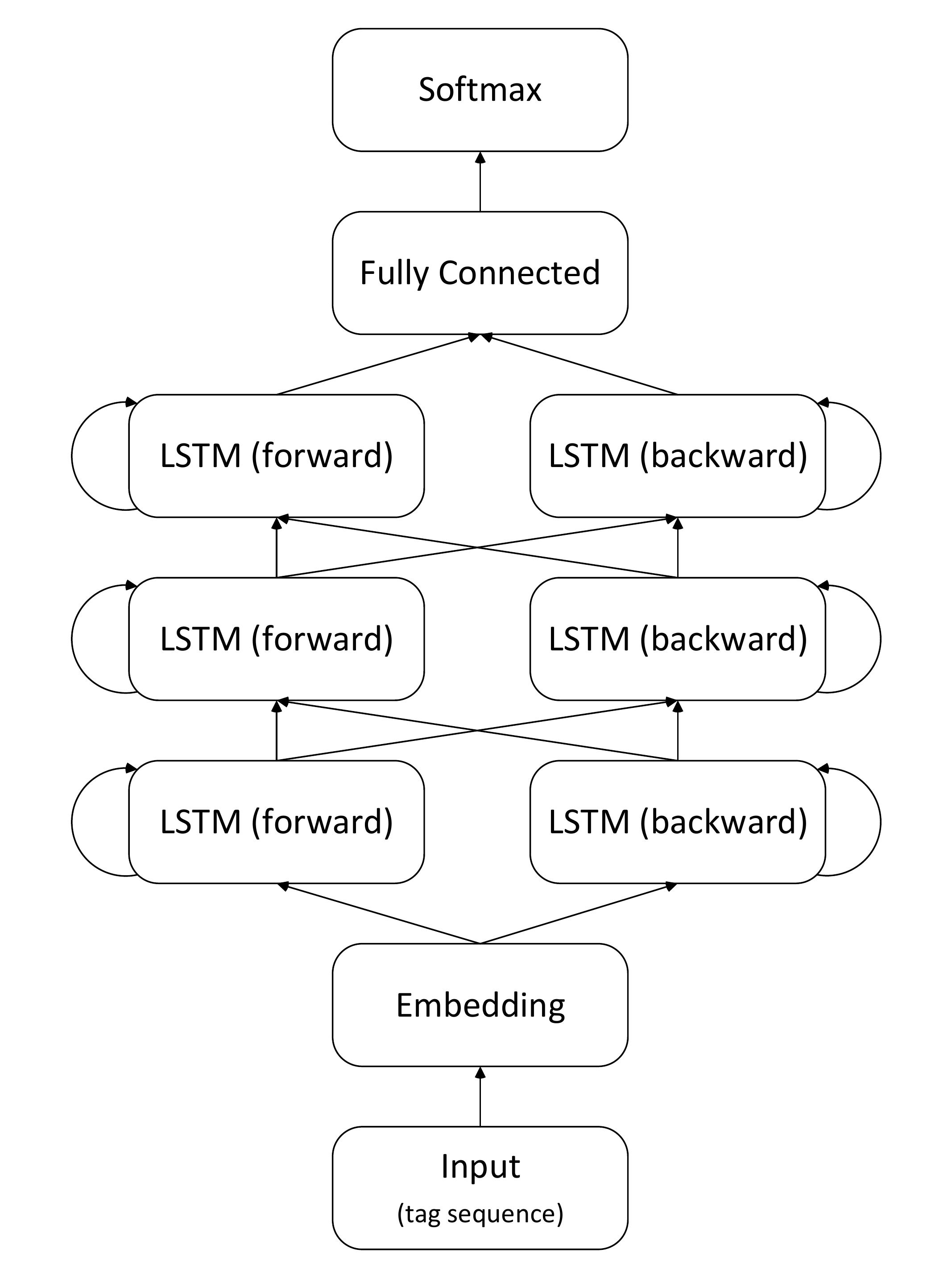}
\caption{Tag prediction neural network architecture.}
\label{fig:network_architecture}
\end{figure}

The embeddings are then fed into several recurrent layers. After the final recurrent layer, the last output of the forward net and the backward net are concatenated and serve as input to a fully-connected layer, followed by a softmax layer, which infers the missing tag in the sentence. The architecture is shown in figure \ref{fig:network_architecture}.

\section{Experiment}
\label{sec:experiment}

In this section we describe the data sets, the specific model architecture settings used in the experiment and the results.

\subsection{Data}
\label{sec:experiment:data}

In order to evaluate the effectiveness of the proposed method, we used two publicly available Polish text corpora - the Polish Sejm Corpus (PSC) \citelanguageresource{ogrodniczuk2012polish} and a subset of the National Corpus of Polish (NCP) \citelanguageresource{przepiorkowski2008towards} \citelanguageresource{przepiorkowski2010recent}. The PSC is a large collection of speeches and interpellations of parliamentary sessions of the Polish government. Although PSC is provided along with automatically extracted morphosyntactic tags, we used the WCRFT tagger \citelanguageresource{radziszewski2013tiered} to extract them from plain PSC text, because of its ability to predict the tags of numerals written as digits. We used default tagger model supplied with the WCRFT software, which was trained on the NCP.  The PSC corpus was split and served as training (99\%) and validation (1\%) data sets. 

As the evaluation data set, we used the publicly available 1-million word subset of the NCP. The main feature of this subset is its manual annotation. Due to the fact that NCP was designed to be representative of the Polish language in general, the corpus consists partly of the PSC, as well as books, newspapers and transcripts of informal speech. Due to its richness, we encountered some tags not found in our annotation of the PSC, which resulted in a total of 321 (8\%) of sentences being removed from the evaluation set.

Even though the NCP corpus could be considered better candidate for training set due to its manual annotations, we chose to train on PSC, which is a much larger corpus, more suitable for training a DNN. We also wanted to make the evaluation criterion more strict, since in this scenario, we also evaluate how well does the model generalize beyond the domain of the Polish parliament language.

\begin{table}[ht]
\centering
\label{tab:datasets}
\begin{tabular}{|c|c|c|c|}
\hline
Data set   & Corpus & Sentences & Words \\
\hline
Train      & PSC    & 521251 & 10684799 \\ 
Validation & PSC    & 5265   & 107511   \\ 
Evaluation & NCP    & 3491   & 71895    \\ 
\hline
\end{tabular}
\caption{Data sets used in the experiment.}
\end{table}

The morphosyntactic information is typically described by a particular tagset. In this work, we adopted the NKJP tagset \citelanguageresource{przepiorkowski2009comparison}, which is being used in the NCP. A major difficulty is the lack of availability of the true morphosyntactic tags for the abbreviations - one of the NKJP tagset assumptions was to manually annotate abbreviations with tags \textit{brev${\colon}$pun} and \textit{brev${\colon}$npun}, which indicate an abbreviation followed/not followed by punctuation, respectively. Since the true tags are available neither in automatic nor manual annotation, we select 34 Polish words which are often abbreviated \ref{tab:abbreviations}, look up their inflected forms in the Polimorf morphological dictionary \citelanguageresource{wolinski2012polimorf},
and gather sentences which contain at least one of these words. As a result, we obtain sentences where an abbreviation might have been possible, and consequently acquire labels (the true morphosyntactic tags) for a supervised machine learning algorithm. We did not consider abbreviations of multiple words in this work. 

The size of each data set is presented in table \ref{tab:datasets}. Sentences containing more than one abbreviation are used repeatedly with different word abbreviated each time in order not to introduce additional complexity into the experiment (which would involve handling more than one \textit{unknown} tag by the network). Also, in order to reduce the training time, we used only sentences of length up to 30 words.

\begin{table}[ht]
\centering
\label{tab:abbreviations}
\begin{tabular}{|c|c|c|c|}
\hline
Abbrev. & Base form & Translation & Freq. [\%] \\
\hline
r & rok & year & 4.29 \\
woj & wojew\'odztwo & voievoidship & 1.44 \\
pkt & punkt & point & 1.32 \\
wyd & wydanie & edition & 0.46 \\
art & artyku\l{} & article & 0.41 \\
ok & oko\l{}o & around & 0.39 \\
godz & godzina & hour & 0.35 \\
poz & pozycja & position & 0.30 \\
jedn & jednolity & uniform & 0.24 \\
z\l{} & z\l{}oty & zloty (PLN) & 0.23 \\
tys & tysiąc & thousand & 0.23 \\
mln & milion & million & 0.15 \\
nr & numer & number & 0.14 \\
prof & profesor & professor & 0.11 \\
\'sw & \'swięty & saint & 0.11 \\
proc & procent & percent & 0.10 \\
ul & ulica & street & 0.08 \\
ks & ksiądz & priest & 0.06 \\
mld & miliard & billion & 0.05 \\
ozn & oznaczenie & mark & 0.05 \\
ha & hektar & hectare & 0.04 \\
dr & doktor & doctor & 0.04 \\
gen & genera\l{} & general & 0.04 \\
ust & ustęp & paragraph & 0.03 \\
l & litr & liter & 0.03 \\
km & kilometr & kilometer & 0.03 \\
m & metr & meter & 0.03 \\
im & imię & name & 0.02 \\
del & delegowany & delegate & 0.02 \\
sygn & sygnatura & signature & 0.02 \\
par & paragraf & paragraph & 0.02 \\
kg & kilogram & kilogram & 0.01 \\
rep & repertorium & repertory & 0.003 \\
cm & centymetr & centimeter & 0.001 \\

\hline
\end{tabular}
\caption{Abbreviations expanded in the experiment, along with their base form and frequency relative to the total word count in PSC corpus.}
\end{table}

\subsection{Experimental setup}
\label{sec:experiment:setup}

We have tested several configurations of the neural network architecture and present the one which achieved the best results. In the training data, we found 950 different morphosyntactic tags (including the special \textit{unknown} tag) and feed them in one-hot-encoded form to the embedding layer, obtaining 32-dimensional real-valued vectors. The bidirectional LSTM layers have 64 hidden units for each direction. After each recurrent layer, we apply dropout \cite{srivastava2014dropout} with a factor of 0.2. After the last recurrent layer, the fully connected layer has 128 hidden units followed by dropout with a factor of 0.5 and ReLU activation \cite{nair2010rectified}. Finally, the softmax layer has 257 output units - this is different from the input layer due to the fact that the abbreviations in our experiment are mostly nouns and are described using only a subset of the original tagset. The last two layers are additionally regularized with weight decay of 0.0005. 

The training was performed using the error backpropagation algorithm \cite{rumelhart1988learning} with cross-entropy criterion and Adam optimizer \cite{kingma2014adam} with the same settings of learning rate and momentum parameters as suggested in the paper. Final parameters of the model were selected from the epoch with minimal validation loss. We used the TensorFlow \cite{tensorflow2015-whitepaper} and Keras \cite{chollet2015keras} frameworks for the experiment.

The correctly inflected abbreviation expansion is obtained by a lookup of the base form and predicted morphological tag in a morphological dictionary - in our case, the Polimorf.

\subsection{Results}
\label{sec:experiment:results}

After 29 epochs of training, the DNN achieved 84.5\% accuracy on the training set, 85.7\% accuracy on the validation set and 74.2\% accuracy on the evaluation set. We hypothesize that the major factor behind the degradation of performance on the evaluation set (in contrast to consistent performance on the train and validation sets) is the difference between tag distribution in both corpora. As mentioned before, the NCP sentences were collected from several different language domains, such as casual language, letters, newspapers and parliament speeches, while the PSC consists entirely of the kind of language encountered in the parliament. It is likely that the tag distributions are different enough to introduce generalization errors, when the classifier was trained on a single domain exclusively.

By contrast, when a simple baseline which predicts the most frequent tag for each abbreviation is assumed, it achieves the following accuracies: training set - 42.8\%, validation set - 42.6\%, evaluation set - 40.3\%. The result obtained by the DNN is significantly stronger and showcases the importance of the abbreviation context, which is not surprising. These results constitute significant evidence that morphosyntactic information is sufficient in many cases.

We would like to showcase and discuss some of the problematic cases in which the DNN failed to predict properly in the evaluation set. First of all, we noticed that about 10\% of accuracy is lost due to confusing singular and plural grammatical number. While most of these are unacceptable, there are a few mistakes which are less severe - e.g. "\textit{2.7 l}" should have been translated as "\textit{2.7 litra}" (\textit{2.7 liter}), but the output was "\textit{2.7 litr\'ow}" (\textit{2.7 liters}). Also, 49 mistakes (about 5\% of accuracy lost) concerned the inflection of the \textit{procent} word, expanded from the \% symbol. It appears that the network learned the rule that in Polish, a noun which has its quantity specified (e.g. \textit{5 centimeters}), is generally inflected with case and number, even though \textit{procent} is an exception to this rule. It follows reason, since the network had no way of knowing which word was used and the exceptional case is less frequent than the general one.

Another category of errors is related to grammatical gender. The Polish words \textit{ulica} (\textit{street}) and \textit{godzina} (\textit{hour}) are feminine, but in 40 cases their abbrevations (\textit{ul.} and \textit{godz.}) were classified as masculine. We suspect that the nature of this error is similar to the previous ones - the network had no way of acquiring prior knowledge about the correct gender of these words.

\section{Conclusions}
\label{sec:conclusions}

We discussed the problem of inflected abbreviation expansion and investigated the viability of expansion based on morphological tags of the abbreviation context combined with lookup in a predefined abbreviation dictionary. We successfully applied a bidirectional LSTM in this task and achieved a reasonable accuracy in an experiment conducted on two Polish corpora. Given the error analysis of our model, we conclude that the morphosyntactic information of the context is not sufficient to deduce the morphosyntactic tags for the expanded abbreviation - although it works in a significant number of cases, prior knowledge about factors such as base form or grammatical gender of the expanded abbreviation is required for correct prediction. We expect that incorporation of this prior knowledge in the model will yield significantly better expansion accuracy.

\section{Bibliographical References}
\label{main:ref}

\bibliographystyle{lrec}
\bibliography{emnlp2017}

\begin{thebibliography}{}

\bibitem[\protect\citename{Abadi \bgroup et al.\egroup
  }2015]{tensorflow2015-whitepaper}
Abadi, M., Agarwal, A., Barham, P., Brevdo, E., Chen, Z., Citro, C., Corrado,
  G.~S., Davis, A., Dean, J., Devin, M., Ghemawat, S., Goodfellow, I., Harp,
  A., Irving, G., Isard, M., Jia, Y., Jozefowicz, R., Kaiser, L., Kudlur, M.,
  Levenberg, J., Man\'{e}, D., Monga, R., Moore, S., Murray, D., Olah, C.,
  Schuster, M., Shlens, J., Steiner, B., Sutskever, I., Talwar, K., Tucker, P.,
  Vanhoucke, V., Vasudevan, V., Vi\'{e}gas, F., Vinyals, O., Warden, P.,
  Wattenberg, M., Wicke, M., Yu, Y., and Zheng, X.
\newblock (2015).
\newblock {TensorFlow}: Large-scale machine learning on heterogeneous systems.
\newblock Software available from tensorflow.org.

\bibitem[\protect\citename{Chollet}2015]{chollet2015keras}
Chollet, F.
\newblock (2015).
\newblock Keras.
\newblock \url{https://github.com/fchollet/keras}.

\bibitem[\protect\citename{Graves and Schmidhuber}2005]{graves2005framewise}
Graves, A. and Schmidhuber, J.
\newblock (2005).
\newblock Framewise phoneme classification with bidirectional {LSTM} and other
  neural network architectures.
\newblock {\em Neural Networks}, 18(5):602--610.

\bibitem[\protect\citename{Hochreiter and Schmidhuber}1997]{hochreiter1997long}
Hochreiter, S. and Schmidhuber, J.
\newblock (1997).
\newblock Long short-term memory.
\newblock {\em Neural computation}, 9(8):1735--1780.

\bibitem[\protect\citename{Huang \bgroup et al.\egroup
  }2015]{huang2015bidirectional}
Huang, Z., Xu, W., and Yu, K.
\newblock (2015).
\newblock Bidirectional {LSTM-CRF} models for sequence tagging.
\newblock {\em arXiv preprint arXiv:1508.01991}.

\bibitem[\protect\citename{Kingma and Ba}2014]{kingma2014adam}
Kingma, D. and Ba, J.
\newblock (2014).
\newblock Adam: A method for stochastic optimization.
\newblock {\em arXiv preprint arXiv:1412.6980}.

\bibitem[\protect\citename{Mikolov \bgroup et al.\egroup
  }2013]{mikolov2013distributed}
Mikolov, T., Sutskever, I., Chen, K., Corrado, G.~S., and Dean, J.
\newblock (2013).
\newblock Distributed representations of words and phrases and their
  compositionality.
\newblock In {\em Advances in neural information processing systems}, pages
  3111--3119.

\bibitem[\protect\citename{Nair and Hinton}2010]{nair2010rectified}
Nair, V. and Hinton, G.~E.
\newblock (2010).
\newblock Rectified linear units improve restricted {B}oltzmann machines.
\newblock In {\em Proceedings of the 27th international conference on machine
  learning (ICML-10)}, pages 807--814.

\bibitem[\protect\citename{Pakhomov}2002]{pakhomov2002semi}
Pakhomov, S.
\newblock (2002).
\newblock Semi-supervised maximum entropy based approach to acronym and
  abbreviation normalization in medical texts.
\newblock In {\em Proceedings of the 40th annual meeting on association for
  computational linguistics}, pages 160--167. Association for Computational
  Linguistics.

\bibitem[\protect\citename{Pennell and Liu}2011]{pennell2011toward}
Pennell, D. and Liu, Y.
\newblock (2011).
\newblock Toward text message normalization: Modeling abbreviation generation.
\newblock In {\em Acoustics, Speech and Signal Processing (ICASSP), 2011 IEEE
  International Conference on}, pages 5364--5367. IEEE.

\bibitem[\protect\citename{Roark and Sproat}2014]{roark2014hippocratic}
Roark, B. and Sproat, R.
\newblock (2014).
\newblock Hippocratic abbreviation expansion.
\newblock In {\em ACL (2)}, pages 364--369.

\bibitem[\protect\citename{Rumelhart \bgroup et al.\egroup
  }1988]{rumelhart1988learning}
Rumelhart, D.~E., Hinton, G.~E., and Williams, R.~J.
\newblock (1988).
\newblock Learning representations by back-propagating errors.
\newblock {\em Cognitive modeling}, 5(3):213--220.

\bibitem[\protect\citename{Sproat}2010]{sproat2010lightly}
Sproat, R.
\newblock (2010).
\newblock Lightly supervised learning of text normalization: {R}ussian number
  names.
\newblock In {\em Spoken Language Technology Workshop (SLT), 2010 IEEE}, pages
  436--441. IEEE.

\bibitem[\protect\citename{Srivastava \bgroup et al.\egroup
  }2014]{srivastava2014dropout}
Srivastava, N., Hinton, G.~E., Krizhevsky, A., Sutskever, I., and
  Salakhutdinov, R.
\newblock (2014).
\newblock Dropout: a simple way to prevent neural networks from overfitting.
\newblock {\em Journal of Machine Learning Research}, 15(1):1929--1958.

\bibitem[\protect\citename{Sundermeyer \bgroup et al.\egroup
  }2012]{sundermeyer2012lstm}
Sundermeyer, M., Schl{\"u}ter, R., and Ney, H.
\newblock (2012).
\newblock {LSTM} neural networks for language modeling.
\newblock In {\em Interspeech}, pages 194--197.

\end{thebibliography}


\begin{thebibliography}{}

\bibitem[\protect\citename{Ogrodniczuk}2012]{ogrodniczuk2012polish}
Ogrodniczuk, M.
\newblock (2012).
\newblock The {P}olish {S}ejm {C}orpus.
\newblock In {\em Proceedings of the Eighth International Conference on
  Language Resources and Evaluation (LREC-2012)}, pages 2219--2223.

\bibitem[\protect\citename{Przepi{\'o}rkowski \bgroup et al.\egroup
  }2008]{przepiorkowski2008towards}
Przepi{\'o}rkowski, A., G{\'o}rski, R.~L., Lewandowska-Tomaszyk, B., and
  Lazinski, M.
\newblock (2008).
\newblock Towards the {N}ational {C}orpus of {P}olish.
\newblock In {\em Proceedings of the Sixth International Conference on Language
  Resources and Evaluation (LREC-2008)}.

\bibitem[\protect\citename{Przepi{\'o}rkowski \bgroup et al.\egroup
  }2010]{przepiorkowski2010recent}
Przepi{\'o}rkowski, A., G{\'o}rski, R.~L., Lazinski, M., and Pezik, P.
\newblock (2010).
\newblock Recent developments in the {N}ational {C}orpus of {P}olish.
\newblock In {\em Proceedings of the Seventh International Conference on
  Language Resources and Evaluation (LREC-2010)}.

\bibitem[\protect\citename{Przepi{\'o}rkowski}2009]{przepiorkowski2009comparison}
Przepi{\'o}rkowski, A.
\newblock (2009).
\newblock A comparison of two morphosyntactic tagsets of {P}olish.
\newblock pages 138--144. Citeseer.

\bibitem[\protect\citename{Radziszewski}2013]{radziszewski2013tiered}
Radziszewski, A.
\newblock (2013).
\newblock A tiered {CRF} tagger for {P}olish.
\newblock In {\em Intelligent tools for building a scientific information
  platform}, pages 215--230. Springer.

\bibitem[\protect\citename{Sproat and Jaitly}2016]{sproat2016rnn}
Sproat, R. and Jaitly, N.
\newblock (2016).
\newblock {RNN} approaches to text normalization: A challenge.
\newblock {\em arXiv preprint arXiv:1611.00068}.

\bibitem[\protect\citename{Wolinski \bgroup et al.\egroup
  }2012]{wolinski2012polimorf}
Wolinski, M., Milkowski, M., Ogrodniczuk, M., and Przepi{\'o}rkowski, A.
\newblock (2012).
\newblock Polimorf: a (not so) new open morphological dictionary for {P}olish.
\newblock In {\em Proceedings of the Eighth International Conference on
  Language Resources and Evaluation (LREC-2012)}, pages 860--864.

\end{thebibliography}

\section{Language Resource References}
\label{lr:ref}
\bibliographystylelanguageresource{lrec}
\bibliographylanguageresource{emnlp2017}

\end{document}